\documentclass[anonymous=false, %
               format=acmsmall, %
               review=false, %
               screen=true, %
               nonacm=true]{acmart}

\usepackage{algorithm}
\usepackage{algorithmic}
\usepackage{url}
\usepackage{mathtools}
\usepackage{hyperref}
\usepackage{listings}
\usepackage{amsthm}
\theoremstyle{definition}

\usepackage{GoMono}

\newcommand{\appropto}{\mathrel{\vcenter{
  \offinterlineskip\halign{\hfil$##$\cr
    \propto\cr\noalign{\kern1pt}\sim\cr\noalign{\kern-1pt}}}}}

\lstdefinelanguage{Golang}%
  {morekeywords=[1]{package,import,func,type,struct,return,defer,panic,%
     recover,select,var,const,iota,},%
   morekeywords=[2]{string,uint,uint8,uint16,uint32,uint64,int,int8,int16,%
     int32,int64,bool,float,float32,float,complex64,complex128,byte,rune,uintptr,%
     error,interface},%
   morekeywords=[3]{map,slice,make,new,nil,len,cap,copy,close,true,false,%
     delete,append,real,imag,complex,chan,},%
   morekeywords=[4]{for,break,continue,range,go,goto,switch,case,fallthrough,if,%
     else,default,},%
   morekeywords=[5]{Println,Printf,Error,Print,},%
   sensitive=true,%
   morecomment=[l]{//},%
   morecomment=[s]{p*}{*/},%
   morestring=[b]',%
   morestring=[b]",%
   morestring=[s]{`}{`},%
}

\acmConference[]{}{}{}

\setcopyright{none}

\title{Bayesian Policy Search for Stochastic Domains}
\author{David Tolpin}
\affiliation{
	\institution{Ben-Gurion University of the Negev}
}
\affiliation{
	\institution{PUB+}
    \country{Israel}
}
\email{david.tolpin@gmail.com}

\author{Yuan Zhou}
\affiliation{
    \institution{University of Oxford}
    \country{United Kingdom}
}
\email{yuan.zhou@cs.ox.ac.uk}

\author{Hongseok Yang}
\affiliation{
	\institution{School of Computing, KAIST}
        \country{South Korea}
}
\email{hongseok00@gmail.com}

\begin{abstract}
\end{abstract}

\begin{document}
\maketitle

\section{Introduction}

AI planning can be cast as inference in probabilistic models,
and probabilistic programming was shown to be capable of policy
search in partially observable domains. Wingate et al.~\cite{WGR+11} introduce
policy search through Markov Chain Monte Carlo.  However, the
work only considers POMDPs with deterministic actions and fixed
initial and final states, as well as deterministic policies.
Handling of non-deterministic POMDPs, uncertain initial and
final states, or stochastic policies are not addressed (and
would involve some form of nesting in the models~\cite{T18}
requiring inference schemes beyond those employed in the paper).
Van de Meent et al.~\cite{MPT+16} adapt black-box variational
inference~\cite{RGB14} to policy search with probabilistic
programs. The use of black-box variational inference facilitates
handling of stochastic POMDPs, in which observations and state
transitions are noisy with known noise distributions. Still, the
handling of stochasticity is tied to a particular inference
algorithm rather than expressed in the model in an
algorithm-agnostic manner. In addition, the application of
black-box variational inference is not strictly Bayesian:
the policy search is performed by optimizing parameters of the
`prior'. As such, the prior just defines the shape of the
variational distribution and the starting point for optimization
rather than encodes prior knowledge about policies.

In this work, we cast policy search in stochastic domains as a
Bayesian inference problem and provide a scheme for encoding
such problems as nested probabilistic programs. We argue that
probabilistic programs for policy search in stochastic domains
should involve \textit{nested conditioning}~\cite{T18}, and
provide an adaption of Lightweight
Metropolis-Hastings~\cite{WSG11} (LMH) for robust inference in
such programs. We apply the proposed scheme to stochastic
domains introduced in~\cite{MPT+16} and show that policies of
similar quality are learned, despite a simpler and more general
inference algorithm. We believe that the proposed variant of LMH
is both novel and, with additional effort, applicable to a wider
class of probabilistic programs with nested conditioning.

\section{Probabilistic Programs for Policy Search}

We consider the problem of inference of a deterministic
parametric policy for an agent in a stochastic partially
observable environment. An environment is described by sets of
actions $A$ and observations $O$. An agent repeatedly performs
an action $a \in A$ and receives an observation $o \in O$ and a
reward. A parametric policy $\pi(\theta): O^{1:t} \rightarrow A$
is a mapping from the history of observations to an action,
fully determined by its parameter $\theta$. The purpose of an
agent is to maximize the sum of rewards $R = \sum_{t=1}^{\infty}
r_t$.  The objective of policy inference is to find $\theta$
that maximizes the expected reward. The interaction between the
agent and the environment is often formalized as a partially
observable Markov decision process (POMDP)~\cite{KLC98}, where
the history of observations summarized by the agent as a belief
state, although formalization as POMDP is not necessary for
policy inference with probabilistic programs. 

A probabilistic program for policy inference is built around a
stochastic simulator $\mathcal{S}(\theta) \rightarrow r$ which
receives policy parameters $\theta$ and returns reward $r$.  The
simulator is usually designed such that it always terminates
after a bounded number of steps, and the reward is also bounded
from both above and below (a reward can be negative).  Since the
simulator is stochastic, $r|\theta$ is a random variable. For
inference, a prior $D_\theta$ is imposed on policy parameters,
and the posterior distribution of $\theta$ is conditioned on an
auxiliary Bernoulli variable, with $\mathbf{1}$ corresponding to
the optimal policy, given the expected reward $\mathbb{E}[r]$.
The conditioning is chosen such that the mode of the posterior
distribution of $\theta$ maximizes the expected reward, and thus
defines the optimal policy for given $\mathcal{S}$.

A common choice for the conditional distribution is~\cite{K07}
\begin{equation}
        p_{c_1}(\mathbf{1}|r) = \exp(r - U_r),
	\label{eqn:exp-r}
\end{equation}
where $U_r$ is an upper bound on the reward.
This results in the following generative model:
\begin{equation}
	\begin{aligned}
		\theta & \sim D_\theta \\
		\mathbf{1} & \sim \mathrm{Bernoulli}(\exp(\mathbb{E}[\mathcal{S}(\theta)] - U_r))
	\end{aligned} \label{eqn:pp-exp-r}
\end{equation}
Model~\eqref{eqn:pp-exp-r} is a model with nested
conditioning~\cite{T18}.  The expectation over outcomes of the
stochastic simulator $\mathcal{S}$ cannot be computed exactly,
and must be approximated by Monte Carlo samples. For
\eqref{eqn:pp-exp-r} an unbiased estimate of $p(\theta)$ cannot
be obtained~\cite{T18}, however a Monte Carlo estimate of $\log
p(\theta)$ is unbiased. Therefore, inference methods which
involve an unbiased estimate of $\log p(\theta)$, such as
stochastic gradient Markov Chain Monte Carlo~\cite{MCF15} or
stochastic variational inference~\cite{HBW+13,RGB14}, can be
efficiently applied to \eqref{eqn:pp-exp-r}, but application of
simpler and more general methods, which do not require computing
gradients, such as importance sampling or Metropolis-Hastings,
faces difficulties of nested Monte Carlo
estimation~\cite{RCY+18}.


An alternative model may use instead 
\begin{equation}
	p_{c_2}(\mathbf{1}|r) = 
        \frac{r - L_r}{U_r - L_r},
	\label{eqn:r}
\end{equation}
where $U_r$ and $L_r$ are an upper and a lower bound on the reward:
\begin{equation}
	\begin{aligned}
		\theta & \sim D_\theta \\
		\mathbf{1} & \sim \mathrm{Bernoulli}\left(\frac {\mathbb{E}[\mathcal{S}(\theta)] - L_r} {U_r - L_r}\right)
	\end{aligned}
	\label{eqn:pp-r}
\end{equation}
The form of nested conditioning in~\eqref{eqn:pp-r} allows
flattening of nested inference~\cite{T18}. Importance sampling
with independent samples of $\theta$ and $r$ (the latter through
a single run of $\mathcal{S}$) gives an unbiased Monte Carlo
estimate of the posterior distribution of $\theta$.  This class
of models does not require nested Monte Carlo estimation or
computing derivatives of probability functions for inference,
and thus potentially applies to a broader range of problems.
Still, importance sampling is of limited use in high-dimensional
models, and a non-trivial parametric policy is likely to depend
on a high-dimensional parameter vector $\theta$.
Metropolis-Hastings, a general Markov Chain Monte Carlo
algorithm that scales better with the dimensionality of the
random space, cannot be applied unmodified. However, an insight
about the structure of the posterior distribution leads to a
simple adaptation of Metropolis-Hastings to~\eqref{eqn:pp-r},
introduced in the next section.

\section{Stochastic Lightweight Metropolis-Hastings}

The conditioning probability of the expected reward
in~\eqref{eqn:pp-exp-r} and~\eqref{eqn:pp-r} can be re-written
in terms of conditioning probabilities of individual rewards.
Indeed, denoting by $p_{\mathcal{S}}(r|\theta)$ the probability
that $\mathcal{S}$ returns $r$ given $\theta$, we get
\footnote{The sum and product are to be replaced by the integral
and the type II geometric integral for the continuous case.}
\begin{equation}
        \label{eqn:pp-comp-r}
	\begin{aligned}
                p_{c_1}(\mathbf{1}|\mathbb{E}(r|\theta)) & = \exp(\mathbb{E}(r|\theta)-U_r) = \prod_r (\exp(r-U_r))^{p_{\mathcal{S}}(r|\theta)} = \prod_r p_{c_1}(\mathbf{1}|r)^{p_\mathcal{S}(r|\theta)},  &&\mbox{ for model~\eqref{eqn:pp-exp-r},} \\
                p_{c_2}(\mathbf{1}|\mathbb{E}(r|\theta)) & = \frac{\mathbb{E}(r|\theta) - L_r}{U_r - L_r} = \sum_r p_{\mathcal{S}}(r|\theta)\left(\frac{r - L_r}{U_r - L_r}\right) = \sum_r {p_\mathcal{S}(r|\theta)}p_{c_2}(\mathbf{1}|r), &&\mbox{ for model~\eqref{eqn:pp-r}.} 
	\end{aligned}
\end{equation}
In words, \eqref{eqn:pp-exp-r} corresponds to conditioning on
conjunction of values of $r$ (all values), and \eqref{eqn:pp-r}
--- on disjunction of values of $r$ (any value), according to
their probabilities.

The $p_{c_2}(\mathbf{1}|\mathbb{E}(r|\theta))$, in particular, takes
the form of the probability of a mixture with a component for
each $r$, component weights $p_{\mathcal{S}}(r|\theta)$ and
conditional probability of a sample given the component
$p_{c_2}(\mathbf{1}|r)$. Componentwise Markov chain Monte Carlo
sampling from a mixture is straightforwardly accomplished by
updating component assignments ($r$) and component parameters
($\theta$). A difficulty arises though because, while $r$ can be
updated given $\theta$ by re-running $\mathcal{S}$, updating
$\theta$ for given $r$ would require running $\mathcal{S}$
backward. This difficulty can be overcome by noting that
stochastic simulator $\mathcal{S}(\theta)$ is a probabilistic
routine itself, and as such can be substituted by a
deterministic procedure $\mathcal{P}(\theta, \tau)$ which takes
an additional parameter $\tau \sim D_\tau$ --- the trace ---
containing all random choices taken during the run of the
simulator. With this substitution, \eqref{eqn:pp-r} takes the
form
\begin{equation}
	\begin{aligned}
		\theta & \sim D_\theta \\
		\mathbf{1} & \sim \mathrm{Bernoulli}\left(\frac {\mathbb{E}_\tau[\mathcal{\mathcal{P}}(\theta, \tau)] - L_r} {U_r - L_r}\right)
	\end{aligned}
	\label{eqn:pp-r-tau}
\end{equation}
Thus, from the calculation similar to the one for the $p_{c_2}$ term in \eqref{eqn:pp-comp-r}, it 
follows that posterior samples $\theta$ of~\eqref{eqn:pp-r-tau} 
can be obtained by first sampling pairs $(\tau,\theta)$ from the joint distribution
\begin{equation}
        \frac{p(\tau) p(\theta) p_{c_2}(\mathbf{1}|\mathcal{P}(\theta,\tau))}{Z}
        = p(\tau) \times \frac{p(\theta) p_{c_2}(\mathbf{1}|\mathcal{P}(\theta,\tau))}{Z_{\tau}}
        = p(\tau) \times \frac{p(\theta) (\mathcal{P}(\theta,\tau) - L_r)}{Z_{\tau} (U_r - L_r)}
\end{equation}
for some normalising constants $Z$ and $Z_{\tau}$, and then selecting the $\theta$ part of each sampled pair. This observation leads to the following sampling algorithm:
\begin{algorithm}[h!]
	\caption{Sampling from~\eqref{eqn:pp-r-tau} with stochastic Metropolis-Hastings}
	\label{alg:smh}
	\begin{algorithmic}[1]
		\LOOP
			\STATE $\tau \sim D_\tau$ \COMMENT{always accept}
			\STATE $\theta' \sim D_\theta$
			\STATE $u \sim \mathrm{Uniform}(0, 1)$
			\IF {$u < \min\left(1, \frac {\mathcal{P}(\theta', \tau) - L_r} {\mathcal{P}(\theta, \tau) - L_r}\right)$}
				\STATE $\theta \gets \theta'$
			\ENDIF
			\STATE Output $\theta$
		\ENDLOOP
	\end{algorithmic}
\end{algorithm}

To prove correctness of Algorithm~\ref{alg:smh},  we consider the joint
distribution of $\tau, \theta$. To sample from this distribution with
componentwise Metropolis-Hastings, change is proposed to either
$\tau$ or $\theta$ on each step. The proposal is accepted
or rejected based on the MH acceptance ratio:
\begin{equation}
	\begin{aligned}
                p_{accept}(\theta'|\theta, \tau) & = \min\left(1, \frac {p(\theta)p(\theta')(\mathcal{P}(\theta', \tau) - L_r)} {p(\theta')p(\theta)(\mathcal{P}(\theta, \tau) - L_r)}\right) = \min\left(1, \frac {\mathcal{P}(\theta', \tau) - L_r} {\mathcal{P}(\theta, \tau) - L_r}\right), \\
		p_{accept}(\tau'|\theta, \tau) & = \min\left(1, \frac {p(\tau)p(\tau')} {p(\tau')p(\tau)}\right) 
                = 1.
	\end{aligned}
\end{equation}
Note that $p_{accept}(\tau'|\theta, \tau) = 1$, which is reflected by ``always accept'' in the algorithm. 


Expected reward need not be estimated for sampling. This allows
to re-write \eqref{eqn:pp-r-tau} as a probabilistic program
without nesting, but with a syntax extension tagging $\tau$ as a
stochastic choice --- a random variable for which the posterior,
rather than the prior distribution is known:
\begin{equation}
	\begin{aligned}
		\theta & \sim D_\theta \\
		\tau & \overset{*}{\sim} D_\tau \\
		\mathbf{1} & \sim \mathrm{Bernoulli}\left(\frac {\mathcal{\mathcal{P}}(\theta, \tau) - L_r} {U_r - L_r}\right)
	\end{aligned}
	\label{eqn:pp-r-tau-flat}
\end{equation}
Correspondingly, in the stochastic variant of lightweight
Metropolis-Hastings~\cite{WSG11}, the only modification required
is always accepting an update to a component marked as
stochastic choice ($\overset{*}{\sim}$). The mode of the
marginal distribution of $\theta$ can be found by simulated
annealing, again, applied to $\theta$ but not to $\tau$. We
believe that stochastic LMH, while introduced here for policy
search in stochastic domains, is applicable to probabilistic
programs with nested conditioning in general, providing a more
scalable alternative to importance sampling. It can also be
extended to gradient-based variants of Markov Chain Monte Carlo
when the gradient is available.

\section{Case Studies}
\label{sec:case-studies}

We re-examine here two of the case studies in~\cite{MPT+16},
Canadian traveller problem and RockSample. We re-state the
probabilistic programs for policy inference according
to~\eqref{eqn:pp-r-tau-flat} and perform posterior inference
using the stochastic variant of lightweight Metropolis-Hastings
with simulated annealing to find the optimal policy as the mode
of the posterior. Based on the original case studies, we
implemented our versions of the probabilistic programs and the
inference algorithm in Anglican~\cite{TMY+16}. The code and data
for the case studies are in
repository~\url{https://bitbucket.org/dtolpin/stochastic-conditioning}.

Apart from cosmetic changes, two modifications of the
probabilistic programs were made:
\begin{enumerate}
	\item stochastic choices where tagged to distinguish between
		them and policy parameters.
	\item the conditioning was changed from~\eqref{eqn:exp-r}
		to~\eqref{eqn:r}.
\end{enumerate}
In both case studies, we report the expected rewards of the
inferred policies, for a range of temperatures from 100 to
0.001. We include temperatures greater than 1 to obtain policy
distributions which are close to policy priors, and show that as
we decrease the temperature, the expected policy reward
increases and then levels out when the policies are sampled from
a close proximity of the mode. For each temperature, we run the
inference for $100\,000$ iterations, and use $10\,000$ episodes
to estimate the expected reward of the inferred policies. The
numerical results are similar to those reported
in~\cite{MPT+16}.

\subsection{Canadian Traveller Problem}

\begin{figure}
    \includegraphics[width=0.9\linewidth]{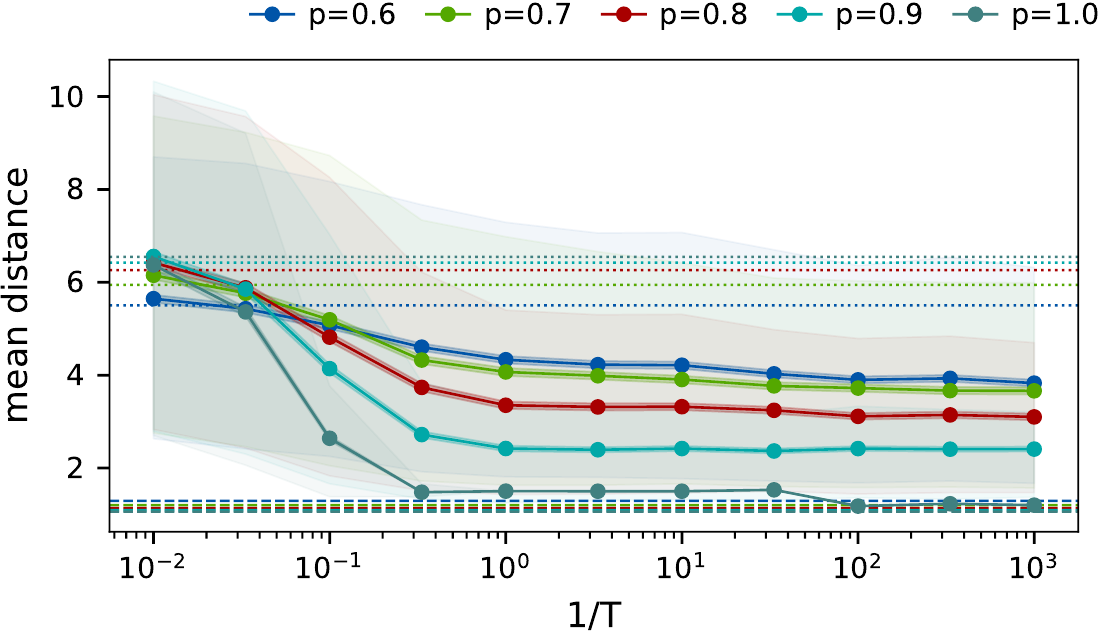}
	\caption{Mean travel distances  for inferred policies in the
	Canadian traveller problem. Dashed and dotted lines are mean travel
	distances by clairvoyant and random agents.} 
    \label{fig:ctp}
\end{figure}

In the Canadian traveller problem (CTP)~\cite{PY89}, an agent
must traverse a graph, in which edges may be missing at random.
The agent knows the length and probability of being open of each
edge.

The agent traverses the graph by depth-first search, and the policy
determines the order in which the edges are selected at each
node.  The reward in CTP is the negative travel distance. The
reward is bounded between the negative sum of lengths of all
edges in the graph (depth-first search visits any edge at
most once) and the shortest path through the graph.
Figure~\ref{fig:ctp} shows the expected travel distances of
inferred policies for a problem instance with 20 nodes and 46
edges, for a range of edge openness probabilities, compared to
the expected travel distances of a random agent and of the
clairvoyant agent (an agent that knows the state of all edges
upfront and travels through the shortest open path).

\begin{figure}
    \includegraphics[width=0.9\linewidth]{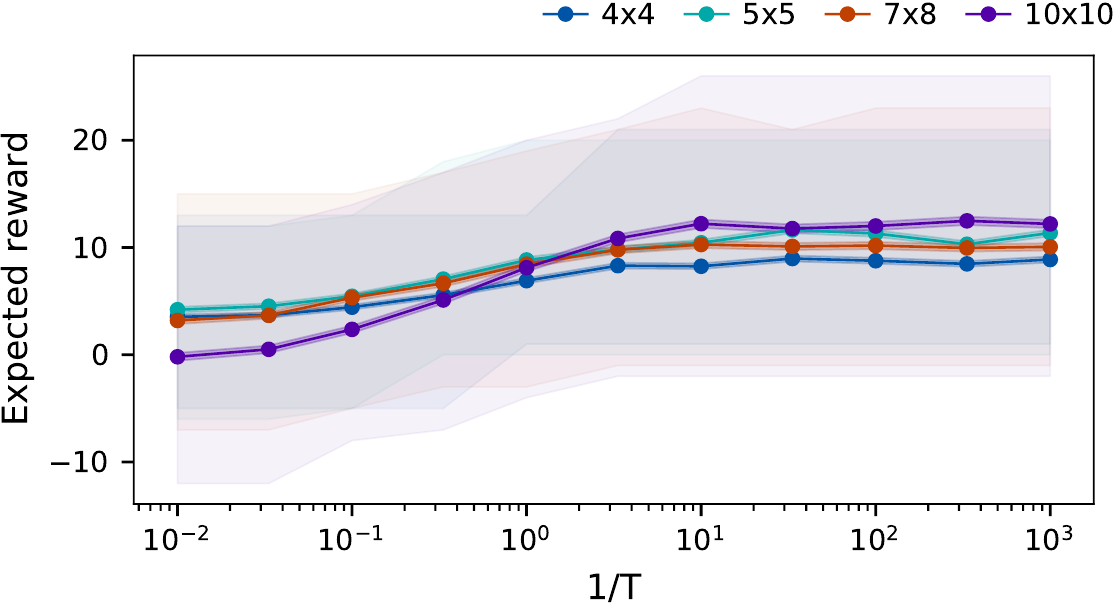}
	\caption{Expected rewards for the RockSample POMDP.}
	\label{fig:rockwalk}
\end{figure}

\subsection{RockSample}

In the RockSample POMDP~\cite{SS04}, a square field $N \times N$
with $M$ rocks is given. A rover is initially located in the
middle of the left edge of the square. Each of the rocks can be
either good or bad; and the rover must traverse the field and
collect samples of good rocks. The rover can sense the quality
of a rock remotely with an accuracy decreasing with the distance
to the rock. The objective is to cross the field fast, while
still sampling as many good rocks as possible. The policy
determines whether to visit or skip a rock based on sensing.
The agent gets a reward of 10 for each good rock, as well as for
reaching the right edge. Every step incurs a cost of -1.  The
reward is not bounded from below, however even a simple
heuristic policy yields a positive reward on the instances used
for evaluation.  Due to that, we use softplus-transformed reward
$\log (\exp(r) + 1)$ for conditioning. Figure~\ref{fig:rockwalk}
shows the expected rewards of inferred policies for four
instances of different sizes and with different numbers of
stones for a range of temperatures.

\section{Discussion}

We cast policy search in stochastic domains as a
Bayesian inference problem and provided a scheme for encoding
such problems as nested probabilistic programs. Bayesian treatment
plays an important role in policy inference; consider,
for example, the Canadian traveller problem. We followed the
model of the original case study and imposed a uniform prior on
the traversal order. Instead, we could impose a prior that
favors directions which are more likely to lead to the goal,
such as directions of the shortest paths from each node to the
goal in the fully unblocked path. \cite{WGR+11} allows to
exploit priors for policy search in deterministic domains, and
our works extends the fully Bayesian approach to stochastic
domains. 

We formalized policy search in stochastic domains as a case of
nested conditioning~\cite{T18}. We proposed a model in which
nested conditioning can be flattened, and non-nested importance
sampling can be used for inference. However, the performance of
importance sampling degrades exponentially with the
dimensionality of the probability space, and policies often 
have many parameters. For example, in the model for the Canadian
traveller problem used here, the number of policy parameters is
proportional to the number of edges in the graph. To facilitate 
faster inference, we introduced a stochastic variant of
lightweight Metropolis-Hastings~\cite{WSG11}, suitable for
Bayesian policy search in stochastic domains, but also in models
with the basic form of stochastic conditioning in general. The
adaptation of LMH to stochastic domains can be applied to other
MCMC variants, e.g. those exploiting the gradient information
where available.

\bibliography{refs}
\bibliographystyle{acm-reference-format}

\end{document}